\documentclass{Interspeech}

\interspeechcameraready

\title{Advancing Automated Speaking Assessment Leveraging Multifaceted Relevance and Grammar Information}

\author[affiliation={1}]{Hao-Chien}{Lu}
\author[affiliation={1}]{Jhen-Ke}{Lin}
\author[affiliation={1}]{Hong-Yun}{Lin}
\author[affiliation={1}]{Chung-Chun}{Wang}
\author[affiliation={1}]{Berlin}{Chen}

\affiliation{Department of Computer Science and Information Engineering}{National Taiwan Normal University}{}
\email{howchien@ntnu.edu.tw, jacob@ntnu.edu.tw, buffett@ntnu.edu.tw, takala@ntnu.edu.tw, berlin@ntnu.edu.tw}

\usepackage{comment}

\begin{document}

\maketitle

\begin{abstract}
    
    Current automated speaking assessment (ASA) systems for use in multi-aspect evaluations often fail to make full use of content relevance, overlooking image or exemplar cues, and employ superficial grammar analysis that lacks detailed error types. This paper ameliorates these deficiencies by introducing two novel enhancements to construct a hybrid scoring model. First, a multifaceted relevance module integrates question and the associated image content, exemplar, and spoken response of an L2 speaker for a comprehensive assessment of content relevance. Second, fine-grained grammar error features are derived using advanced grammar error correction (GEC) and detailed annotation to identify specific error categories. Experiments and ablation studies demonstrate that these components significantly improve the evaluation of content relevance, language use, and overall ASA performance, highlighting the benefits of using richer, more nuanced feature sets for holistic speaking assessment.
\end{abstract}


\section{Introduction}

Automated speaking assessment (ASA) has garnered significant attention in the fields of computer-assisted language learning (CALL) and large-scale language testing. ASA systems provide real-time, detailed feedback, which is beneficial for many second language (L2) learners who seek to improve their spoken proficiency. Moreover, ASA systems can reduce the reliance on human scoring.

Early studies in ASA \cite{vowel-paper,POS,offtopic-content,GEC_BANNO} primarily focused on a single aspect, including delivery (e.g., pronunciation, fluency, and intonation), language use (e.g., part-of-speech (POS) tags, syntactic dependencies (DEP), GEC, and morphology), and content (e.g., topic relevance and coherence). However, given the intricate and complex nature of spoken responses, ASA should adopt a multi-aspect design to capture the interplay between delivery, language use, and content. It is anticipated that a comprehensive framework would better reflect real-world communicative competence.

To date, only a few studies have explored multi-aspect designs in ASA \cite{ETS-sys,fast-task,SAMAD,wranksim}. These approaches typically combine handcrafted features (such as pitch, duration, POS tags, and syntactic dependencies) and self-supervised learning (SSL) features (such as wav2vec and BERT) to predict scores for delivery, language use, and content via late fusion of the aspect embeddings. Notably, the SAMAD model \cite{SAMAD} advanced this area by incorporating cross-aspect attention mechanisms for feature integration.

However, existing research often defines content solely in terms of the relevance between the question and the response, overlooking cross-modal topic relevance. For example, studies such as \cite{SAMAD,wranksim} fail to incorporate image relevance in image description tasks. Additionally, for language use, most studies rely on low-level features for grammar error detection \cite{pos-gec}, without addressing fine-grained grammatical error types.

In this paper, we propose a hybrid model for ASA that enhances content evaluation through a novel multifaceted relevance module and refines language use analysis using fine-grained grammar error features. Through extensive experiments and ablation studies, we demonstrate how these components overcome key limitations in current ASA systems and significantly improve assessment accuracy.

Prior work often adopts a narrow definition of content relevance and employs superficial grammatical analysis. Our multifaceted relevance module expands content evaluation by integrating exemplar, image, and response information to capture both topic and image relevance. For language use, we introduce grammar features derived from advanced GEC models with fine-grained error annotations, enabling the identification of specific error types. These features are fused using cross-aspect attention, alongside delivery features, to form a unified hybrid model. Our previous approach (dubbed SAMAD) \cite{SAMAD} informs our implementation of cross-aspect attention for feature fusion. Experimental results confirm that the proposed enhancements lead to improved performance in both aspect-level and overall proficiency evaluations.

In summary, this paper presents three main contributions:
\begin{enumerate}
\item A multifaceted relevance module integrating exemplar, image, and question-response data for comprehensive content assessment.

\item Detailed grammar error features via advanced GEC  and annotation from the SERRANT system for nuanced language use evaluation.

\item An enhanced hybrid model effectively integrating these novel features to improve overall ASA performance.
\end{enumerate}

To our knowledge, this work is the first to enhance multi-aspect ASA by concurrently introducing a multifaceted relevance module and fine-grained grammar error features, demonstrating their combined efficacy in an integrated model.

\section{Methodology}

\begin{figure*}[!t]
  \centering
  \includegraphics[width=0.7\linewidth]{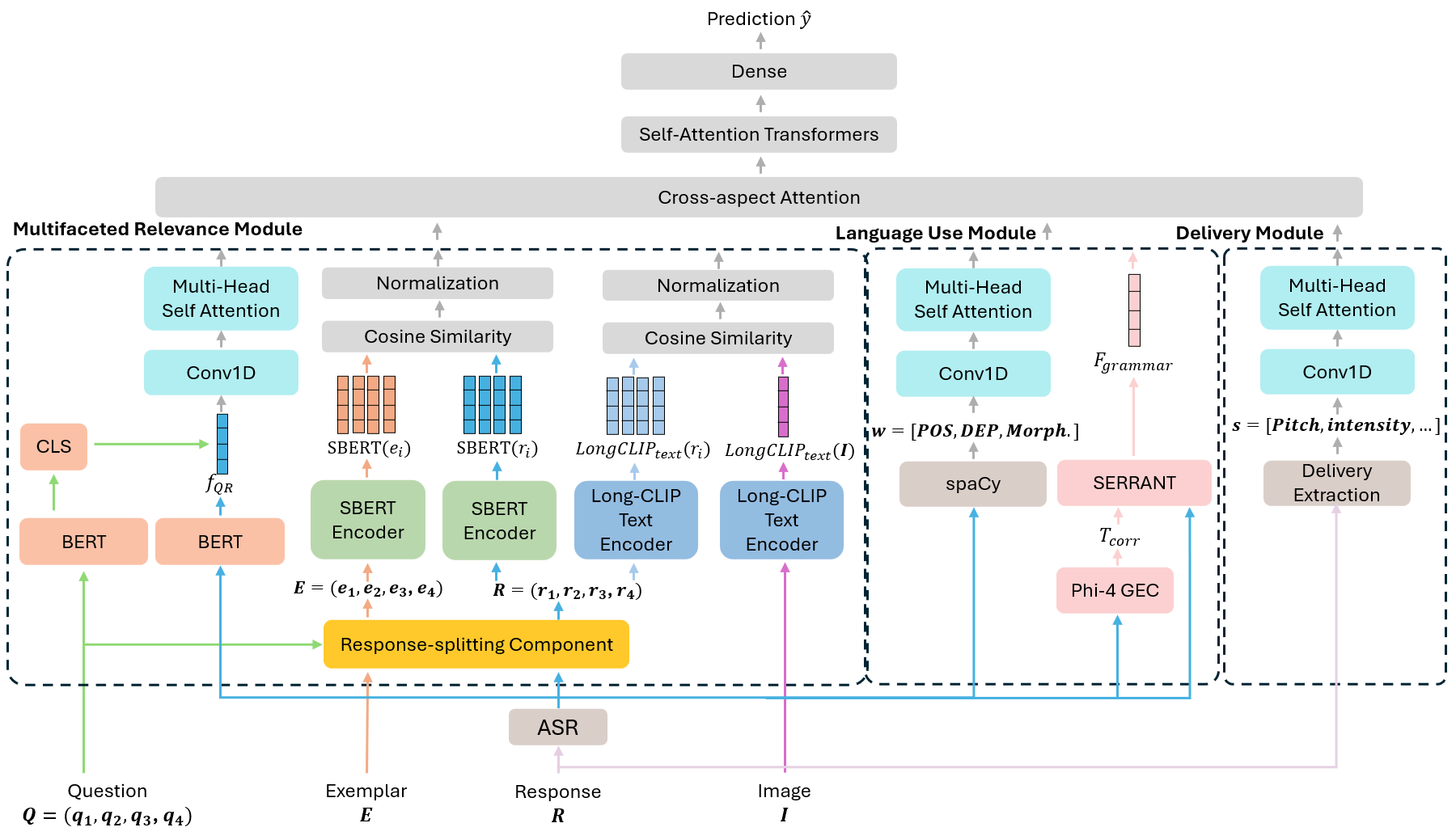}
  \caption{Our proposed system}
  \label{fig:overll-arch}
\end{figure*}

Figure~\ref{fig:overll-arch} illustrates our proposed system, which evaluates spoken proficiency through three key aspects: relevance, language use, and delivery. To enhance content evaluation, we propose a multifaceted relevance module (Section 2.1) that integrates information from the image, exemplar, question, and response to comprehensively assess relevance. For language use, we introduce grammar error features (Section 2.2) to capture detailed grammatical issues. Finally, we present a hybrid model (Section 2.3) that integrates all these features to produce a unified and robust assessment.

\subsection{Multifaceted Relevance Module}
The module consists of three components: (1) a response-splitting component that segments the response according to each question, (2) an exemplar-response relevance component that evaluates the relevance between exemplars and responses, and (3) an image-response relevance component that assesses the relevance between images and responses. This design accounts for cases where certain questions may not be directly related to the image. By cohesively integrating exemplars, images, and responses, the module enables the extraction of additional information beyond what is depicted in the image alone.

\textbf{The Response-splitting Component}
This component aims to transform a respondent's overall response into individual answers aligned with each question $q_j$ in a given set of questions $Q = \{q_1, q_2, \ldots, q_k\}$. Initially, an ASR system transcribes the respondent's audio input $A$. This raw text output from the ASR is then processed by LLaMA-3.1-8B-Instruct, denoted as $M_{split}$, using a carefully designed prompt $P_{prompt}$ to generate structured outputs $R_{split} = \{r_1, r_2, \ldots, r_k\}$, where $r_j$ is the response segment corresponding to question $q_j$:
\begin{align}
R_{split} = M_{split}(\text{ASR}(A), Q, P_{prompt}).
\end{align}
The prompt begins with: "You are a helpful assistant. You are given several questions and a single consolidated answer. The answer contains the information needed to respond to each question. Your task: For each question below, please provide the relevant text from the answer. Only output what is found in the answer..." (remaining lines specify the format instructions). The same process is applied to an exemplar text $E_{text}$ to obtain corresponding split exemplar answers $E_{split} = \{e_1, e_2, \ldots, e_k\}$.

\textbf{The Exemplar-Response Relevance Component}
This component evaluates the semantic relevance between an exemplar segment $e_i \in E_{split}$ and its corresponding response segment $r_i \in R_{split}$. First, the response-splitting component extracts the answers corresponding to each question $q_i$ from both the exemplar (yielding $e_i$) and the response (yielding $r_i$). Then, the Sentence-Transformer (SBERT) \cite{sentence-transformer} is used to compute sentence embeddings for these segments. The semantic similarity $sim_{ER,i}$ between the paired segments is then measured using cosine similarity between their respective embeddings:
\begin{align}
sim_{ER,i} = \text{sim}(\text{SBERT}(e_i), 
\text{SBERT}(r_i)).
\end{align}
Finally, we normalize these similarity scores $sim_{ER,i}$ to obtain standardized scores $S_{ER,i}$ for all questions, mapping them to the range $[0.01, 1]$ (assigning $S_{ER,i} = 0$ if $r_i$ represents no response). This normalization ensures consistent and comparable relevance evaluation.

\textbf{The Image-Response Relevance Component}
This component measures the relevance between the image $I$ and each question-aligned response segment $r_i \in R_{split}$. First, the response-splitting component is used to extract the question-aligned answer $r_i$ from the overall response. We then use the Long-CLIP \cite{Long-CLIP} image encoder for image embeddings and the Long-CLIP text encoder for text embeddings from the response segment. Compared to CLIP \cite{CLIP}, Long-CLIP supports long-text inputs, effectively addressing CLIP's input length limitations. We compute the cosine similarity $sim_{IR,i}$ between the image and text embeddings to evaluate their relevance:
\begin{align}
sim_{IR,i} = \text{sim}(\text{LongCLIP}_{\text{image}}(I), \text{LongCLIP}_{\text{text}}(r_i)).
\end{align}
The same normalization procedure described in Section 2.1.2 is applied to $sim_{IR,i}$ to yield the normalized score $S_{IR,i}$.

\textbf{The Question-Response Relevance Component}
This component assesses the topical alignment between the question $Q$ and the response $R$. A BERT-based feature extractor processes $Q$ and $R$ to generate a sequence of contextual embeddings $f_{QR} = (f_{QR,1}, f_{QR,2}, \ldots, f_{QR,M})$, where $M$ is the number of tokens in $R$. Each embedding $f_{QR,i}$ in this sequence is formed by concatenating the [CLS] token embedding derived from $\text{BERT(Q)}$ with the $i$-th token embedding derived from $\text{BERT(R)}$:
\begin{align}
f_{QR,i} = \text{concat}(\text{BERT}(Q)_{[CLS]},\text{BERT}(R)_i).
\end{align}
This sequence $f_{QR}$ (where each $f_{QR,i} \in \mathbb{R}^{d_{QR}}$) is subsequently processed by further layers in the model, as illustrated in Figure~\ref{fig:overll-arch}.

\subsection{Grammar Error Features}
We first use an ASR system to transcribe the respondent’s audio input $A$. Then, we employ Microsoft’s Phi-4 model as the GEC component, denoted $M_{GEC}$. This model processes the ASR output ($T_{raw}$) to produce its corrected version $T_{corr}$:
\begin{align}
T_{corr} = M_{GEC}(\text{ASR}(A)).
\end{align}
The Phi-4 model is fine-tuned using the Speak \& Improve dataset \cite{sicorpus25}, with further enhancements via few-shot learning by providing three in-context examples from the training data.

To annotate grammatical errors, we adopt SERRANT \cite{SERRANT}, which extends the common ERRANT framework \cite{ERRANT} by introducing SErCl types that provide more informative and fine-grained error categories—particularly in handling “OTHER” and morphology-related errors. SERRANT compares the raw text $T_{raw}$ and the corrected text $T_{corr}$ to identify a set of errors. Let $E_{types} = \{type_1, type_2, \ldots, type_K\}$ be the $K$ distinct grammatical error types defined by SERRANT. For each error type $type_k \in E_{types}$, we determine its count, $c_k$, based on the SERRANT analysis of $T_{raw}$ and $T_{corr}$.

Finally, we construct a feature vector $F_{grammar}$. This vector represents the frequency of each error type, normalized by the total word count $N_w(T_{raw})$ in the original raw text response. The normalized frequency $freq_k$ for each error type $type_k$ is calculated as:
\begin{align}
freq_k = \frac{c_k}{N_w(T_{raw})}.
\end{align}
The resulting grammar error feature vector $F_{grammar}$ is then composed of these frequencies:
\begin{align}
F_{grammar} = [freq_1, freq_2, \ldots, freq_K].
\end{align}
This vector $F_{grammar} \in \mathbb{R}^K$ serves as the input for the language use module.

\subsection{Hybrid Model}
Our approach is based on a multi-aspect evaluation framework, which concurrently analyzes three critical dimensions of a learner’s speech: content, language use, and delivery.

\textbf{Content Module}
The content module evaluates how well the response aligns with the given task. We employ our multifaceted relevance module as the core of this component. Compared to previous methods, our module offers a more comprehensive evaluation of relevance by integrating diverse information sources—such as the question, exemplar, image, and response—thereby enabling a more nuanced understanding of content relevance.

\textbf{Language Use Module}
The language use module focuses on assessing both grammatical accuracy and syntactic structure. For syntactic features, we extract information using spaCy, including POS tags, DEP, and morphological features, collectively referred to as our syntax features. These syntax features are concatenated with the grammar error features (described in Section 2.2) to form multiple one-hot vectors representing different dimensions of language use.

\textbf{Delivery Module}
The delivery module focuses extracts a comprehensive set of acoustic and fluency features. These features are derived from established methodologies and include pitch contours, intensity variations, pause and silence durations, speech rate, and articulation characteristics.

\textbf{Overall Module}
This module integrates features from three distinct perspectives: content, language use, and delivery. 

The processing pipeline for these input features includes the following key components:

\begin{itemize}
\item Cross-Aspect Attention component: To enable dynamic interaction and information fusion across different aspects, a cross-aspect attention mechanism is employed (e.g., Delivery→Content and Language Use→Content). This allows the model to weigh the importance of features from one aspect when processing another.

\item Self-Attention Transformers and Prediction: The outputs from the cross-aspect attention components are concatenated and further processed by a self-attention Transformer layer to gather temporal information before being passed to fully-connected layers for the final proficiency score prediction.

\end{itemize}

\section{Experiments and Results}

\subsection{Data}

\begin{figure}[!ht]
  \centering
  \includegraphics[width=\linewidth]{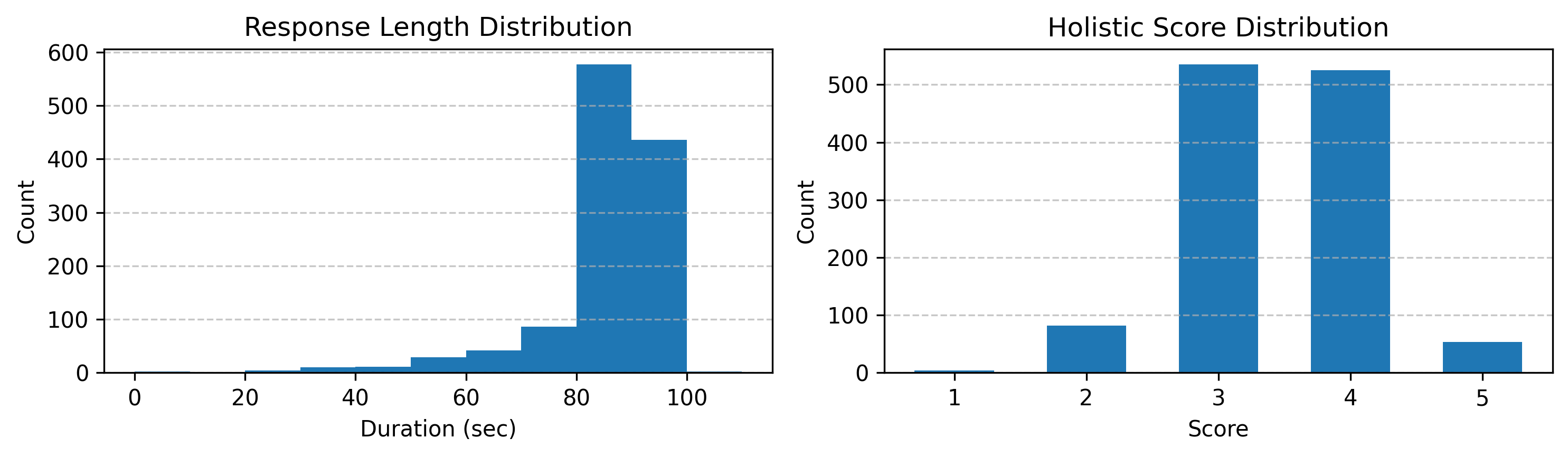}
  \caption{The left figure shows response length distribution in the GEPT corpus, while the other displays holistic score distribution.}
  \label{fig:holistic-score}
\end{figure}

\begin{figure}[!ht]
  \centering
  \includegraphics[width=\linewidth]{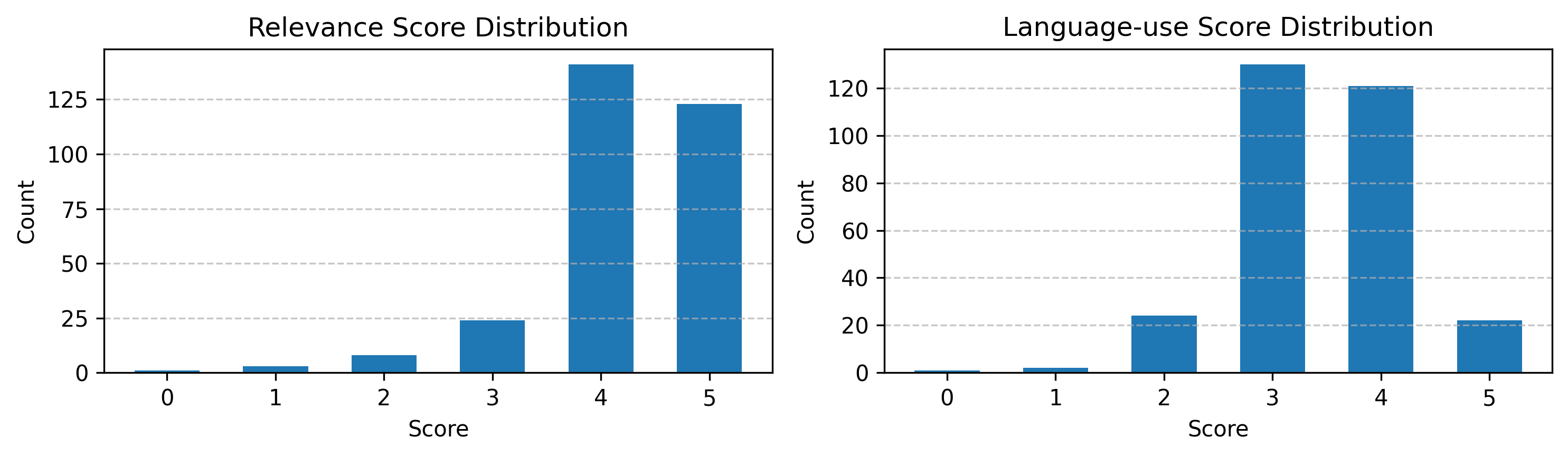}
  \caption{The left figure shows relevance score, while the other displays language use score}
  \label{fig:subscore}
\end{figure}

This study is based on a private corpus provided by the Language Training and Testing Center (LTTC), specifically collected from image-based descriptive tasks at the intermediate level of the General English Proficiency Test (GEPT), a major English assessment in Taiwan. The corpus contains 1,199 responses, evenly distributed across four question sets, each provided by a unique test taker.

Each response was scored by two professional raters on a scale from 1 to 5. The final score was obtained by averaging the two scores and rounding down to the nearest integer.

To evaluate generalization to unseen prompts, one of the four sets was randomly selected as the “unknown” test set. The remaining responses were split into training, development, and “known” test sets in an 80/10/10 ratio.

In addition to holistic scores, one question set also includes relevance and language use scores. A separate dataset was created from these scores, following the same 80/10/10 ratio.

The distributions of response length, holistic, relevance, and language use scores are shown in Figure~\ref{fig:holistic-score} and Figure~\ref{fig:subscore}, respectively.

\subsection{Experimental Setup}
We employ WhisperX \cite{whisperX} for ASR, noted for its accuracy with long-form audio. Input features comprise content features (256-D for question-response, 4-D for exemplar-response, and 4-D for image-response), language use features (247-D for syntax and 265-D for grammar), and delivery features (14-D speech vectors).

To capture temporal dependencies or relationships across different time steps within each sequence, the question-response (256-D), syntax (247-D), and speech (14-D) features are each processed through 1D convolutional layers, then a 3-layer multi-head self-attention transformer (4 heads, 256 hidden dimension). In parallel, the exemplar-response (4-D), image-response (4-D), and grammar (265-D) features are directly projected into 256-D representations.

All 256-D representations resulting from these distinct processes are then fed into a Transformer-based cross-aspect attention. Then, processed by a final self-attention block. Finally, projection layers with residual connections generate the prediction. The model is trained for 32 epochs using AdamW (batch size 32, learning rate $1\times10^5$).

\subsection{Grammar component Analysis}

\begin{table}[th]
  \caption{Compare word error rate (WER) with different GEC models}
  \label{tab:wer-gec}
  \centering
  \begin{tabular}{ lc }
    \toprule
     & \textbf{WER(\%)}\\
    \midrule
    S\&I 2025 baseline             & 17.77\\
    \midrule
    zero-shot Phi-4      & 15.25\\
    few-shot Phi-4       & \textbf{14.44} \\
    \bottomrule
  \end{tabular}
  
  \caption{Experiment results on language use score}
  \label{tab:error tagging}
  \centering
  \begin{tabular}{ lc }
    \toprule
     & \textbf{Accuracy}\\
    \midrule
    w/o normalized\\
    \midrule
    ERRANT             & 0.413\\
    SERRANT            & 0.413 \\
    \midrule
    normalized\\
    \midrule
    ERRANT      & 0.652\\
    SERRANT       & \textbf{0.674} \\
    \bottomrule
  \end{tabular} 
\end{table}

Table~\ref{tab:wer-gec} displays the performance metrics of our GEC component in the Speak \& Improve Challenge 2025. We outperformed baseline models by achieving a significant reduction in error rates. In Table~\ref{tab:error tagging}, we use the language use score to evaluate the accuracy using ERRANT and SERRANT. The results indicate that with normalization, SERRANT has more informative error annotations, leading to improved performance in evaluating the language use.

\subsection{Multi-factor Relevance Comparison}
\begin{table}[th]
  \caption{Experiment results on relevance score}
  \label{tab:multifaceted}
  \centering
  \begin{tabular}{ lc }
    \toprule
     & \textbf{Accuracy}\\
    \midrule
    w/o response-splitting component\\
    \midrule
    only exemplar-response             & 0.475\\
    only image-response      & 0.500\\
    exemplar-response + image-response      & 0.525 \\
    \midrule
    with response-splitting component \\
    \midrule
    only exemplar-response             & 0.600\\
    only image-response      & 0.500\\
    exemplar-response + image-response      & \textbf{0.675} \\
    \bottomrule
  \end{tabular} 
\end{table}

We use the relevance score to evaluate the effectiveness of our proposed multifaceted relevance module. In Table~\ref{tab:multifaceted}, we conduct an ablation study on the various combinations of components and demonstrate the effectiveness of the response-splitting and integrated approach. 

\subsection{Hybrid Comparison}

\begin{table}[!ht]
  \centering
  \caption{Experiment results on holistic score. 'Accuracy (Acc.)' means the prediction must be the same as the score, while 'Binary Accuracy (Bin Acc.)' categorizes scores into two groups: those below 4 (e.g., scores 1, 2, 3) and those 4 and above (e.g., scores 4, 5).}
  \label{tab:holistic-result}
  \begin{tabular}{lcccc}
    \toprule
    Model & \multicolumn{2}{c}{Known Test} & \multicolumn{2}{c}{Unknown Test} \\
    \cmidrule(lr){2-3} \cmidrule(lr){4-5}
     & Acc. & Bin Acc. & Acc. & Bin Acc. \\
    \midrule
    SAMAD \cite{SAMAD}     & 0.567 & 0.711 & 0.687 & 0.753 \\
    \midrule
    ours                & \textbf{0.700} & \textbf{0.789} & \textbf{0.717} & \textbf{0.797} \\
    w/o Grammar          & 0.700 & 0.789 & 0.693 & 0.780 \\
    w/o Multifaceted    & 0.678 & 0.756 & 0.630 & 0.713 \\
    \bottomrule
  \end{tabular}
\end{table}

To establish a strong baseline for evaluating our proposed system, we first replicated the SAMAD model [7]. Table~\ref{tab:holistic-result} presents the performance of this replicated SAMAD model alongside our proposed hybrid models. In this table, we conduct an ablation study on the various combinations of our hybrid model's components, highlighting the impact of each proposed component on overall performance.

The experiments show that both the grammar features and the multifaceted relevance module contribute to the hybrid model. However, the multifaceted relevance module contributes more significantly than the grammar features.

\section{Conclusion}

We enhanced multi-aspect ASA by introducing two key components. Firstly, a multifaceted relevance module improves content evaluation by integrating questions, exemplars, image contents and spoken responses. Secondly, we incorporated detailed grammar error features, derived from advanced GEC and SERRANT annotations, to provide a more nuanced analysis of language use. Integrating these components into an enhanced hybrid model significantly improved ASA performance, yielding a more comprehensive and nuanced assessment system.

\textbf{Limitations and future work.} The efficacy of the crucial response-splitting component is inherently tied to the specific design of its prompt, potentially limiting its accuracy when encountering diverse or unanticipated spoken response styles. Future work will involve investigating and developing more robust techniques for response-splitting.

\section{Acknowledgement}
This work was supported by the Language Training and Testing Center (LTTC), Taiwan. Any findings and implications in the paper do not necessarily reflect those of the sponsor.

\bibliographystyle{IEEEtran}
\bibliography{mybib}

\begin{thebibliography}{10}
\providecommand{\url}[1]{#1}
\csname url@samestyle\endcsname
\providecommand{\newblock}{\relax}
\providecommand{\bibinfo}[2]{#2}
\providecommand{\BIBentrySTDinterwordspacing}{\spaceskip=0pt\relax}
\providecommand{\BIBentryALTinterwordstretchfactor}{4}
\providecommand{\BIBentryALTinterwordspacing}{\spaceskip=\fontdimen2\font plus
\BIBentryALTinterwordstretchfactor\fontdimen3\font minus \fontdimen4\font\relax}
\providecommand{\BIBforeignlanguage}[2]{{%
\expandafter\ifx\csname l@#1\endcsname\relax
\typeout{** WARNING: IEEEtran.bst: No hyphenation pattern has been}%
\typeout{** loaded for the language `#1'. Using the pattern for}%
\typeout{** the default language instead.}%
\else
\language=\csname l@#1\endcsname
\fi
#2}}
\providecommand{\BIBdecl}{\relax}
\BIBdecl

\bibitem{vowel-paper}
L.~Chen, K.~Evanini, and X.~Sun, ``Assessment of non-native speech using vowel space characteristics,'' in \emph{2010 IEEE Spoken Language Technology Workshop}, 2010, pp. 139--144.

\bibitem{POS}
\BIBentryALTinterwordspacing
S.~Bhat and S.-Y. Yoon, ``Automatic assessment of syntactic complexity for spontaneous speech scoring,'' \emph{Speech Communication}, vol.~67, pp. 42--57, 2015. [Online]. Available: \url{https://www.sciencedirect.com/science/article/pii/S0167639314000715}
\BIBentrySTDinterwordspacing

\bibitem{offtopic-content}
C.~M. Lee, S.-Y. Yoon, X.~Wang, M.~Mulholland, I.~Choi, and K.~Evanini, ``Off-topic spoken response detection using siamese convolutional neural networks,'' in \emph{Interspeech 2017}, 2017, pp. 1427--1431.

\bibitem{GEC_BANNO}
\BIBentryALTinterwordspacing
S.~Bannò and M.~Matassoni, ``Back to grammar: Using grammatical error correction to automatically assess l2 speaking proficiency,'' \emph{Speech Communication}, vol. 157, p. 103025, 2024. [Online]. Available: \url{https://www.sciencedirect.com/science/article/pii/S0167639323001590}
\BIBentrySTDinterwordspacing

\bibitem{ETS-sys}
Y.~Qian, P.~Lange, K.~Evanini, R.~Pugh, R.~Ubale, M.~Mulholland, and X.~Wang, ``Neural approaches to automated speech scoring of monologue and dialogue responses,'' in \emph{ICASSP 2019 - 2019 IEEE International Conference on Acoustics, Speech and Signal Processing (ICASSP)}, 2019, pp. 8112--8116.

\bibitem{fast-task}
B.~Lin and L.~Wang, ``Fast task-specific adaptation in spoken language assessment with meta-learning,'' in \emph{ICASSP 2022 - 2022 IEEE International Conference on Acoustics, Speech and Signal Processing (ICASSP)}, 2022, pp. 7257--7261.

\bibitem{SAMAD}
W.-H. Peng, S.~Chen, and B.~Chen, ``Enhancing automatic speech assessment leveraging heterogeneous features and soft labels for ordinal classification,'' in \emph{2024 IEEE Spoken Language Technology Workshop (SLT)}, 2024, pp. 945--952.

\bibitem{wranksim}
C.-W. Wu and B.~Chen, ``Optimizing automatic speech assessment: W-ranksim regularization and hybrid feature fusion strategies,'' in \emph{Interspeech 2024}, 2024, pp. 4004--4008.

\bibitem{pos-gec}
A.~Tezcan, V.~Hoste, and L.~Macken, ``A neural network architecture for detecting grammatical errors in statistical machine translation,'' \emph{The Prague Bulletin of Mathematical Linguistics}, vol. 108, 06 2017.

\bibitem{sentence-transformer}
\BIBentryALTinterwordspacing
N.~Reimers and I.~Gurevych, ``Sentence-{BERT}: Sentence embeddings using {S}iamese {BERT}-networks,'' in \emph{Proceedings of the 2019 Conference on Empirical Methods in Natural Language Processing and the 9th International Joint Conference on Natural Language Processing (EMNLP-IJCNLP)}, K.~Inui, J.~Jiang, V.~Ng, and X.~Wan, Eds.\hskip 1em plus 0.5em minus 0.4em\relax Hong Kong, China: Association for Computational Linguistics, Nov. 2019, pp. 3982--3992. [Online]. Available: \url{https://aclanthology.org/D19-1410/}
\BIBentrySTDinterwordspacing

\bibitem{Long-CLIP}
\BIBentryALTinterwordspacing
B.~Zhang, P.~Zhang, X.~Dong, Y.~Zang, and J.~Wang, ``Long-clip: Unlocking the long-text capability of clip,'' in \emph{Computer Vision – ECCV 2024: 18th European Conference, Milan, Italy, September 29–October 4, 2024, Proceedings, Part LI}.\hskip 1em plus 0.5em minus 0.4em\relax Berlin, Heidelberg: Springer-Verlag, 2024, p. 310–325. [Online]. Available: \url{https://doi.org/10.1007/978-3-031-72983-6_18}
\BIBentrySTDinterwordspacing

\bibitem{CLIP}
\BIBentryALTinterwordspacing
A.~Radford, J.~W. Kim, C.~Hallacy, A.~Ramesh, G.~Goh, S.~Agarwal, G.~Sastry, A.~Askell, P.~Mishkin, J.~Clark, G.~Krueger, and I.~Sutskever, ``Learning transferable visual models from natural language supervision,'' in \emph{Proceedings of the 38th International Conference on Machine Learning}, ser. Proceedings of Machine Learning Research, M.~Meila and T.~Zhang, Eds., vol. 139.\hskip 1em plus 0.5em minus 0.4em\relax PMLR, 18--24 Jul 2021, pp. 8748--8763. [Online]. Available: \url{https://proceedings.mlr.press/v139/radford21a.html}
\BIBentrySTDinterwordspacing

\bibitem{sicorpus25}
\BIBentryALTinterwordspacing
K.~Knill, D.~Nicholls, M.~J. Gales, M.~Qian, and P.~Stroinski, ``{The Speak \& Improve Corpus 2025: an L2 English Speech Corpus for Language Assessment and Feedback},'' 2025. [Online]. Available: \url{https://doi.org/10.17863/CAM.114333}
\BIBentrySTDinterwordspacing

\bibitem{SERRANT}
\BIBentryALTinterwordspacing
L.~Choshen, M.~Oren, D.~Nikolaev, and O.~Abend, ``Serrant: a syntactic classifier for english grammatical error types,'' 2021. [Online]. Available: \url{https://arxiv.org/abs/2104.02310}
\BIBentrySTDinterwordspacing

\bibitem{ERRANT}
\BIBentryALTinterwordspacing
C.~Bryant, M.~Felice, and T.~Briscoe, ``Automatic annotation and evaluation of error types for grammatical error correction,'' in \emph{Proceedings of the 55th Annual Meeting of the Association for Computational Linguistics (Volume 1: Long Papers)}, R.~Barzilay and M.-Y. Kan, Eds.\hskip 1em plus 0.5em minus 0.4em\relax Vancouver, Canada: Association for Computational Linguistics, Jul. 2017, pp. 793--805. [Online]. Available: \url{https://aclanthology.org/P17-1074/}
\BIBentrySTDinterwordspacing

\bibitem{whisperX}
M.~Bain, J.~Huh, T.~Han, and A.~Zisserman, ``Whisperx: Time-accurate speech transcription of long-form audio,'' in \emph{Interspeech 2023}, 2023, pp. 4489--4493.

\end{thebibliography}

\end{document}